# Aggregated Channels Network for Real-Time Pedestrian Detection

Farzin Ghorban[1,2], Javier Marín[3], Yu Su[2], Alessandro Colombo[2], Anton Kummert[1]
[1]Universität Wuppertal, [2]Delphi Deutschland, [3]Massachusetts Institute of Technology

## ABSTRACT

Convolutional neural networks (CNNs) have demonstrated their superiority in numerous computer vision tasks, yet their computational cost results prohibitive for many real-time applications such as pedestrian detection which is usually performed on low-consumption hardware. In order to alleviate this drawback, most strategies focus on using a two-stage cascade approach. Essentially, in the first stage a fast method generates a significant but reduced amount of high quality proposals that later, in the second stage, are evaluated by the CNN. In this work, we propose a novel detection pipeline that further benefits from the two-stage cascade strategy. More concretely, the enriched and subsequently compressed features used in the first stage are reused as the CNN input. As a consequence, a simpler network architecture, adapted for such small input sizes, allows to achieve real-time performance and obtain results close to the state-of-the-art while running significantly faster without the use of GPU. In particular, considering that the proposed pipeline runs in frame rate, the achieved performance is highly competitive. We furthermore demonstrate that the proposed pipeline on itself can serve as an effective proposal generator.

## 1. INTRODUCTION

Pedestrian detection is a canonical case of object detection with a significant relevance to a number of applications in robotics, surveillance and advanced driver assistance systems amongst other fields. Due to the diversity of the appearance of pedestrians including clothing, pose, occlusion, as well as background clutter, pedestrian detection has been considered as one of the most challenging tasks of image understanding for decades.

For solving the task of pedestrian detection, a wide spectrum of approaches have been used in the recent years [3, 10, 15]. As surveyed in [3], existing approaches addressing pedestrian detection can be categorized into three different families. Decision forests [2, 11, 25, 37, 44], deformable parts models (DPM) [12, 13, 27, 29, 39], and deep networks [1, 4, 19, 22, 38, 41]. Within these families, thanks to the recent efforts in the research and the advances in parallel computing hardware, deep networks have exhibited impressive accuracy gains compared to the other approaches [5, 18, 21, 33, 35]. Specifically, in pedestrian detection, CNNs have practically monopolized the state-of-the-art. However, to achieve satisfying performance, the commonly used CNN-based pipelines rely on architecturally massive CNNs which therefore are computationally very expensive. This shortcoming largely prevents their deployment in commercial applications such as advanced driver assistance systems (ADAS) in which the use of costly GPU is not yet practical. This motivates us to design a CNN-based detector with a much smaller network size that can achieve a high performance.

In this work, we propose a very competitive arrangement of a fast detector and a CNN for pedestrian detection, that is both very accurate and runs in real-time using only a single CPU core. To achieve this, we reduce substantially the total number of multiplications caused by the CNN. In particular, we use the fast ACF detector [7] to generate the proposals and introduce a new pipeline which enables the detector to share the ACF feature planes with the CNN. To evaluate the compressed proposals, we design a simple and small network ($2 \cdot 10^5$ parameters), ACNet (Aggregated Channels Network), that performs at a low computational cost. The proposed network is trained from scratch, without initializing the weights from pre-trained models and without using additional data for the training. Our implementation is based on two open source codes, namely, ACF detector provided by Piotr Dollár [7] and MatConvNet [36].

### 1.1 Related Work

Most CNN-based object detection approaches arrange a strong CNN at the end of a proposal generation method. In [1, 19, 45], authors rely on three different instances of the well-known channel features detectors family [2, 8, 11, 25, 44] to generate proposals. In all the cases, the best results are reported when using pre-trained CNN models [1, 21, 33] or a larger dataset such as Imagenet to initially train their network and adapt/fine-tune it to the target dataset. A similar method is also proposed in [4], where authors further extend the number of channel features used in the cascade and incorporate features coming from a CNN. Here the main difference is that the extracted features are arranged in an efficient manner based on their complexity. Region based CNNs (R-CNN) methods [16, 17, 22, 30] have recently dragged the attention of researchers due to their success in object recognition tasks. One of the most remarkable examples, Fast R-CNN [16], shares convolutions across proposals which significantly decreases the computational cost of the network. In such a manner, most of the computation time is expended to generate the proposals. Ren et al. in [30] go further and design a region proposal network that efficiently shares the features with the Fast R-CNN. Other extensions such as [22] modify the original Fast R-CNN network architecture to incorporate a scale aware scheme aiming to improve the performance among different scales (small/large). In [40, 43], the lower layers from a pre-trained network are used to generate features and boosted forests are built on top of that. It is also shown that after a certain depth of the network, the performance gets worse. Here, the CNN features are computed at the beginning, differing from previous works, and then the boosted forest acts as the main classifier. Close to our idea, few methods in the literature have investigated the use of handcraft features, such as HOG [6] or CSS [32], as the CNN input [26, 28, 42], and even other color spaces, YUV instead of RGB, [23, 31]. Differing from them, our method is built on top of aggregated channels and the use of a reduced CNN, which efficiently reuses the computed features in the first stage, to cope with this shrunk input. Although the aforementioned works constitute a step forward in our research field, in most of the cases their massive computational cost makes them impractical for real-time applications on embedded hardware.

### 1.2 Contribution

As far as we know, this is the first work that explores the possibility of reusing aggregated channel features computed in the proposals generation stage. In addition, the proposed CNN is the first one of its kind handling such small input sizes (e.g. $16 \times 8$ pixels for each channel) for pedestrian detection. This novel approach is able to achieve real-time performance without using any GPU computation. In fact, when compared with the state-of-the-arts on Caltech and Kitti datasets, even with those methods running on GPU, ACNet obtains high detection accuracy at a very competitive run-time using only a single CPU core. Furthermore, this paper is the first one exploring the use of ResNet50 as the last component of a cascade. Readers will also find in this paper a comprehensive summary and comparison of most relevant works in pedestrian detection focusing on real-time performance.

## 2. AGGREGATED CHANNELS NETWORK

The idea of cascading a detector with a CNN is based on decreasing the number of candidates in an image. The detector collects proposals and then a strong CNN re-scores them. By modifying the operating point of the detector, one can increase or decrease the number of proposals, and therefore, the cost caused by the CNN and the detector. In the following sections we describe how our two-stage cascade is designed and the implementation details.

### 2.1 Proposal generation

In the first stage, we make use of the well-known ACF detector [11]. ACF has proven to be one of the fastest existing detectors with source code available online [7]. ACF is the descendant of the ICF detector [8, 9] with the main difference that instead of using Haar like features, it divides channels into $4 \times 4$ blocks, where the pixels in each block are summed/aggregated. Every value in these aggregated channels represents a feature, more explicitly, features are single pixel lookups. Furthermore, ACF uses 10 channels: normalized gradient magnitude (1 channel), histogram of oriented gradients (6 channels), and LUV color (3 channels). With a detection window size of $64 \times 32$ and the $4 \times 4$ aggregation, ACF creates proposals of the size $16 \times 8 \times 10$ (1280 features). For the detection, 2048 trees with depth 2 are applied to every detection window in a constant soft cascade manner. In order to make our results easily reproducible, we use the pre-trained model provided in [7].

### 2.2 Proposal evaluation

In the past, a variety of architectures has been proposed and tested successfully in pedestrian detection, e.g. CifarNet [20], AlexNet [21] and, more recently, VGG [33]. Although, such architectures were not initially addressed for pedestrian detection, authors have retrained/adapted them for this specific task. Previous works in the literature performing a sliding window approach, run first the detector for reducing the number of proposals and then the CNN on the corresponding raw RGB windows. In our pipeline, the CNN evaluates the same windows the detector examines during classification. This has the advantage that we do not have to revisit the original RGB image and resize the candidates, instead, our CNN uses the same candidate windows produced in the feature pyramid by the detector. This is an important gain compared to existing approaches, since resizing is a costly procedure, especially when it comes to practical real-life pedestrian detection systems, where embedded hardware is involved.

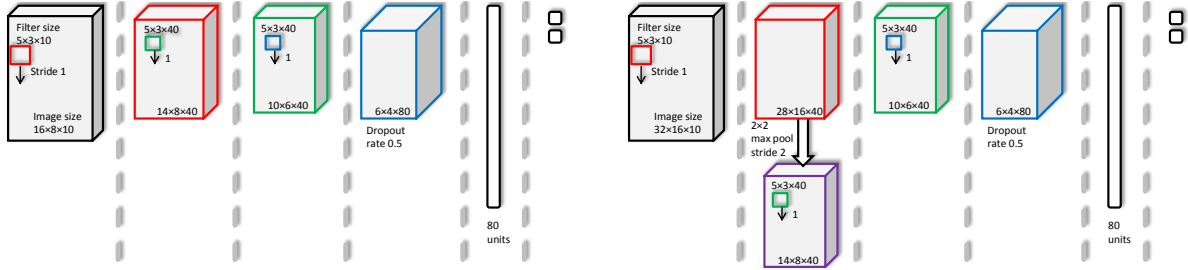

Figure 1. Left: ACNet, total number of multiplications in the convolution layers is $\sim 3 \cdot 10^6$. Right: ACNet+, total number of multiplications in the convolution layers is $\sim 5 \cdot 10^6$.

By sharing the feature scale pyramid between detector and CNN, the detector offers candidates with a resolution of $16 \times 8 \times 10$, since the ACF detector uses a window detection size of $64 \times 32$ and a $4 \times 4$ aggregation. In a sense, ACF channels can be considered as the first layer of the CNN. This is in contrast to the commonly end-to-end features learning approach where CNNs are very successful. In our pipeline, ACF representations are used as separated, non-trained features. To evaluate the offered proposals, we design a small CNN which is architecturally derived from the well known CifarNet [20]. We modify CifarNet, which was successfully tested for the case of pedestrian detection [19], to fit the predefined input size. The CifarNet, when compared to other networks used for pedestrian detection, is one of the smallest ($10^5$ parameters) CNNs [19]. Originally designed to solve the CIFAR-10 classification problem, it was trained using ten object classes, $(5 + 1) \cdot 10^5$ color images with a resolution of $32 \times 32$ pixels [20]. By taking a closer look on the traditional approaches [1, 19], we can immediately see that, the computational cost caused by the max pooling and non-linearity layers can be ignored when compared to that of the convolution layers. For instance, for a raw RGB image patch of the size $64 \times 32 \times 3$, CifarNet needs to do $\sim 25$ millions of multiplications to get a classification score. We are focusing on increasing the speed by decreasing the overall number of multiplications. An intuitive way to achieve that is either to decrease the overall number of weights in the CNN or to decrease the dimension of the input image. In ACNet, the latter way is used and the number of multiplications is reduced to $\sim 3$ millions.

In order to combine our CNN with ACF, we change the network architecture of the CifarNet in different aspects. First, inspired by [34] we remove the subsampling layers, so that the network can still have a sufficient depth. Beside having a positive impact of the learned filters, by making them more insensitive to translation, subsampling layers (or stride 2 convolution layers) are mainly used to successively bringing down the resolution of the input to one dimension at FC-layer, where the neurons have the same receptive field as the input resolution and the classification task can be done. By removing them, dimension reduction can be done in a more gentle manner through the convolution filters $f: \mathbb{R}^{H \times W \times C} \to \mathbb{R}^{H' \times W' \times C'}$, where $H^{(\prime)}, W^{(\prime)}$ and $C^{(\prime)}$ refer to the height, width and depth of the input and the output. Given $N$ filters of the dimension $H'' \times W'' \times C''$, without the use of zero-padding and with stride of $1 \times 1$, the convolution operation results in on output volume of the dimension $H' = H - H'' + 1$, $W' = W - W'' + 1$, and $C' = N$. Second, to avoid disproportionately using of zero-padding, we change the filter sizes in the CifarNet to $C_1 = C_2 = C_3 = 5 \times 3$ (where $C_i$ denotes the $i^{th}$ convolution layer). Third, we replace the regularization layers (i.e. contrast normalization layers) by a computationally much cheaper dropout layer. Finally, we replace the non-linearity of fully connected layers by sigmoid instead of ReLU, and change the objective function of the CNN from softmax and cross-entropy to L2 loss defined as $E = \frac{1}{2} \sum_{i=1}^{N} (t_i - O_i)^2$ where $E$ is the squared error, $N$ is the number of the input samples, $t_i$ is the label of $i^{th}$ sample and $O_i$ is its corresponding network output. Since there is no pre-processing of the input and normalization of the intermediate convolution results, the sigmoid non-linearity and L2 loss lead to a smoother learning behavior and a slightly better performance, as we have observed in our initial tests. Figure 1 illustrates the architecture of ACNet.

### 2.3 Implementation details

It is well known that for CNNs the number of training samples is quite important to reach good performance. In our experiments, we make use of two of the largest and most challenging datasets publicly available in the literature, namely Caltech and Kitti datasets.

**Datasets:** The Caltech dataset and its associated benchmark [3, 10] are frequently used in the field of pedestrian detection. The frames in this dataset are captured from a vehicle driving through urban traffic. All frames are labeled with bounding boxes of pedestrian instances. The standard training set (every $30^{th}$ frame is used) contains 4250 frames with nearly 2000 annotated pedestrians and the test set covers 4024 frames with 1000 pedestrians. Since all the frames in the videos are fully annotated, the size of the training set can be easily increased by re-sampling the videos. Motivated by [25, 44], we enlarge the amount of the training data by almost a factor of 10 (this is called Caltech10× in the literature). We are using Caltech10× for all our experiments.

The Kitti dataset [14] consists of images captured from a car traversing German streets. Its training set contains 4445 pedestrians, 4024 taller than 40 pixels, over 7481 frames, and its test set 7518 frames.

**Validation:** In Caltech dataset we follow the strategy proposed in [10]. The training set is divided into six different video-sequences. We use the first five video-sequences for training and the last one for validation. We follow a 4-fold cross-validation strategy. The training set is split into four subsets, in a way that the number of pedestrians is approximately the same (ignored pedestrians are not taken into account) in each subset. In total, four detectors are evaluated using three subsets for training and the remaining set for validation. In order to train ACNet, we make use of the ACF detector to collect samples.

| Positives | Negatives | MR[%] (BN) | MR[%] (AN) |
|---|---|---|---|
| GT | IoU < 0.5 | 41.9 | 50.1 |
| GT | IoU < 0.3 | 45.0 | 51.7 |
| GT, IoU > 0.5 | IoU < 0.5 | ***41.4*** | 46.9 |
| GT, IoU > 0.5 | IoU < 0.3 | 42.7 | 47.7 |

Table 1. Performance of our ACNet on Caltech validation set trained while using different training sets. MR: log-average miss-rate (lower is better). BN: ACNet is applied before non-maximum suppression. AN: ACNet is applied after non-maximum suppression. GT: ground truth bounding boxes. IoU: Intersection-over-Union of the bounding boxes and the ground truths.

We train the CNN with different training sets by varying the Intersection over Union (IoU) metric, which measures the matching quality and is defined as $IoU = area(GT \cap derections)/area(GT \cup detections)$, where GT denotes the ground truth bounding boxes labeling the pedestrian and detection is the bounding box output of the detector. Table 1 and Table 2 show our results on the Caltech and Kitti validation sets, respectively. In both experiments, ACNet is directly applied after the ACF detector, before and after non-maximum suppression. Since our ACNet does not have any subsampling layer, it behaves very sensitive to translation. In order to match the translation in the positive candidates offered by the detector, we use the detector to collect positive jittered samples. For every ground truth we take the detection with the highest score and an IoU > 0.5. In Table 1 and Table 2 show that using jittered positive training data helps improving the performance.

| Positives | Negatives | AP[%] (BN) | AP[%] (AN) |
|---|---|---|---|
| GT | IoU < 0.5 | 70.4 | 65.9 |
| GT | IoU < 0.3 | 69.2 | 65.1 |
| GT, IoU > 0.5 | IoU < 0.5 | ***75.6*** | 68.9 |
| GT, IoU > 0.5 | IoU < 0.3 | 68.0 | 67.2 |

Table 2. Performance of our ACNet on the Kitti validation set while using different training sets. AP: average precision (higher is better). BN: ACNet is applied before non-maximum suppression. AN: ACNet is applied after non-maximum suppression. GT: ground truth bounding boxes. IoU: Intersection-over-Union of the bounding boxes and the ground truths.

In our experiments, we found out that the ACNet benefits from having more candidates around the pedestrian, which is why its performance increases when it is applied before non-maximum suppression (NMS). We tried different operating points for our baseline detector, by calibrating the weights of the decision trees [47], and concluded that restricting the number of candidates per image to 40 on average seems to provide enough number of positive proposals per pedestrian. More concretely, the ACF detector achieves at this operating point (without non-maximum suppression) a true positive rate of more than 92% on the validation set.

**Training:** We train the network by following the bootstrapping strategy. We start with stage 0 (false positive samples are collected using the ACF detector), train the network, use it to collect more negative samples and continue training with additionally collected negative samples. We continue bootstrapping until the network saturates and performance does not improve. Even by using Caltech10×, pedestrians are rare in the dataset. The number of the collected negative samples is much higher than the number of positive samples existing. In order to handle this imbalance, we restrict the number of collected negative samples by setting at FPPI (false positive per image) = 1 a threshold for the network. At this operating point, only one false positive per image on average is collected. We have also tried other techniques such as using just a subset of all false positives for each training batch with different ratios ($ratio = |negative|/|positive|$, where $|negative|$ and $|positive|$ denote the number of positive and negative samples in the training set, respectively) but observed that employing any restriction of this kind has in the best case no effect on the performance. The network is randomly initialized with Gaussian with mean zero and standard deviation 0.2 and trained from scratch. We use stochastic gradient descent with a batch size of 1000. The learning rate starts from 0.02 and is divided by 10 when the error plateaus. We use a weight decay of 0.0005 and a momentum of 0.9.

## 2.4 Extending the pipeline

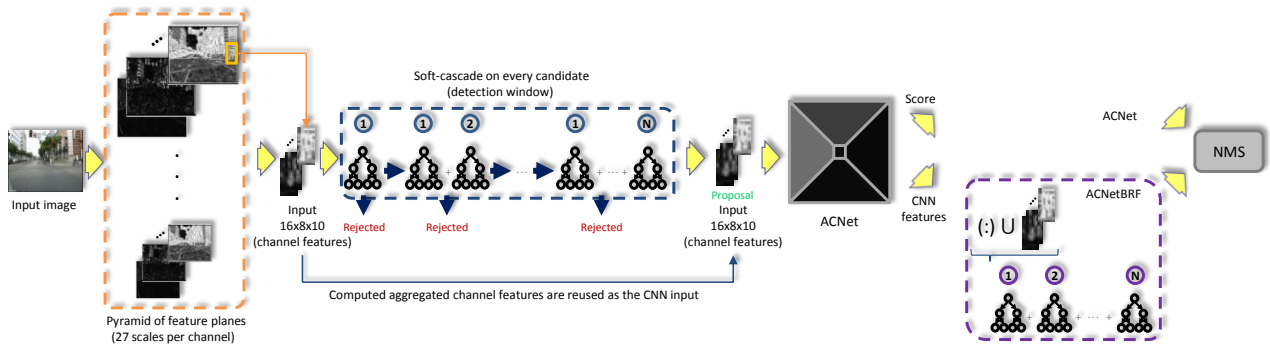

Figure 2. ACNet pipeline, primarily aimed to reduce the overall consumed time. The pipeline can be divided in three processing steps. First, the feature pyramid, created by the detector in order to detect at different scales (orange dashed box). Second, the constant soft-cascade, where every content of the detection window goes through (in blue dashed box). In the cascade process, as soon as the score of the content falls below a predefined threshold, the detection window will be rejected. And third, our ACNet (top right) and ACNetBRF (bottom right in violet dashed box), where every detection window that passes the soft cascade will be seen as a proposal and, therefore, evaluated to get a new score.

**ACNet+:** To see how far we can benefit from the use of aggregated channels, we explore the use of ACF+ [25], which is also open source and with code available online [7]. Compared to ACF, the ACF+ detector uses more data for the training, 4096 depth 5 trees for the binary classification and $2 \times 2$ aggregated feature planes. The detection window size is $64 \times 32$ as before, which creates with the $2 \times 2$ aggregation candidates of the size $32 \times 16 \times 10$, thus it results in 5120 features per each candidate. In order to combine the CNN with ACF+ detector, we add a subsampling layer directly after the first convolution layer. Other layers remain as described in Section 2.2. Because of the different aggregation parameter, it is not possible to use the same CNN for both detectors. We train the CNN following the same procedure in described in Section 2.3. Again, using jittered positive samples with IoU > 0.5 and negative samples with IoU < 0.5 provide the best performance. During our experiments we found out that using negative samples with an IoU below 0.5 leads to lower localization errors at lower FPPI range. We know that, at higher FPPI range the ability of discriminating foreground and background plays a mainly important role. Although the split into positive and negative samples using 0.5 as the IoU threshold seems to be a harsh criterion, it has proven to have no negative effect in the discriminating capability at higher FPPI value.

**ACNetBRF/ACNet+BRF:** Inspired by [43, 40], we enrich the ACF/ACF+ proposals with additional convolution features taken from ACNet/ACNet+, respectively. In order to perform this in a memory friendly way, we vectorize and concatenate the input (ACF/ACF+ planes) and all three intermediate convolution outputs of our ACNet/ACNet+. By first vectorizing these features, we can skip the use of zero-padding. In the case of ACNetBRF, we create a feature vector of the size 1 × 10080 by first vectorizing 16 × 8 × 10, 14 × 8 × 40, 10 × 6 × 40, and 6 × 4 × 80 features from the input, first, second and third convolution layers, respectively and then concatenating them. In case of ACNet+, we take the response of the first convolution layer after the pooling layer, so that only the size of the detector's proposal (32 × 16 × 10) differs, thus the size of the feature vector becomes 1 × 13920. The enriched proposal vector is then fed into a boosted random forest (BRF) consisting of maximum 4096 decision trees with maximum depth of 5 and trained via discrete Adaboost. Furthermore, a random subset of the size $1/16 \cdot |features|$ is used to create every non-leaf node of the trees, where $|features|$ denotes the size of the feature vector. To be more precise, the proposals passing the first stage will be re-scored by the BRF, where BRF does not have any rejection trace or constant threshold. We refer to this extension as ACNetBRF/ACNet+BRF. Both pipelines, ACNet/ACNet+ and ACNetBRF/ACNetBRF+, are depicted in Figure 2. ACNetBRF/ACNet+BRF are also applied before NMS. We train the BRF extension using the same bootstrapped training set used for training the underlying ACNet/ACNet+ network and do not further bootstrap.

**ACNet+BRFResNet:** In [45], it is shown that the performance of deep CNNs tightly depends on the quality of the proposal generator. Considering the effectiveness of ACNet and its variants, they can be used as an effective proposal generator. To validate this point, we extend the pipeline by using a pre-trained ResNet50 model [18]. ResNet50 gets cropped RGB image patches of the size 224 × 224 × 3. However, such large patches are computationally expensive. We modify the network by removing the three top layers (fully connected layer, average pooling and ReLU layer) coming directly after the last sum layer, we then randomly initialize two fully connected layers with 512 and 2 units, respectively and fine-tune the network using [36]. This allows us to use the canonical 128 × 64 × 3 size for the pedestrian template. For the training we collect initially proposals using the ACNet+BRF pipeline. We first train the two new fully connected layers until the error gets plateaus and then train all layers. We perform bootstrapping (using a threshold on FPPI=1, as in Section 2.3) using the new extended pipeline to which we refer as ACNet+BRFResNet and continue until no further improvement is possible. Note, that ResNet50 is attached to the ACNet+BRF pipeline after NMS, where the number of the proposals is further reduced through NMS.

## 2.5 Purging proposals

Motivated by [24, 29], all the detections, coming from the first stage that do not have a plausible relationship between their vertical position and scale are removed. For that, we make use of the same assumptions and the ground truths available in the training data. The main goal of this technique is to reduce the number of candidates the CNN examines and if possible increase the overall detection rate. In some cases, certain hard negatives may be located on unusual locations in the image. Figure 3 shows how the pipeline clearly benefits from this technique. In Kitti we did not use such technique due to the fluctuations in the image resolution (height ∈ [370, 376] and width ∈ [1242, 1224]).

## 3. EXPERIMENTS

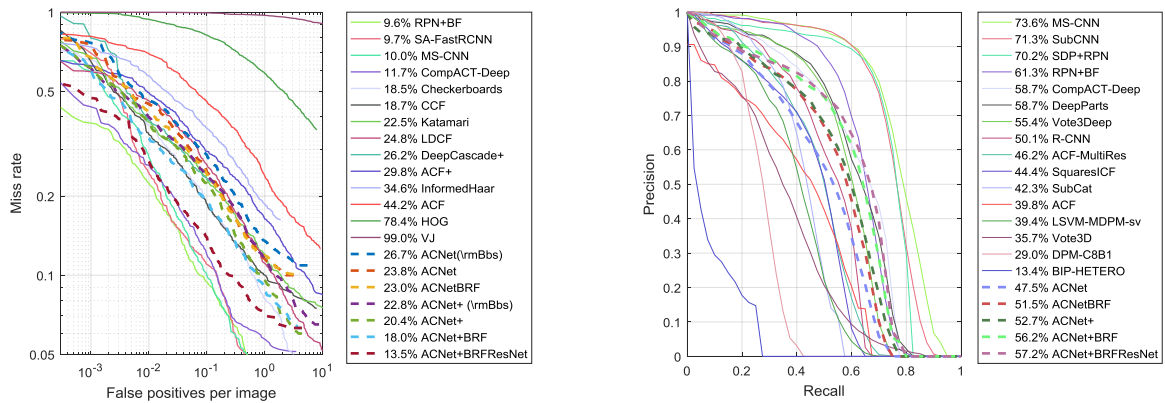

Figure 3. Left: comparison to state-of-the-art on Caltech test set (reasonable setting). Methods are compared using the log-average miss rate metric (MR) measured in the range [$10^{-2}$, $10^0$] false positive per image, lower is better. \rmBbs refers to excluding the purging method (see Section 2.5). Right: comparison to state-of-the-art on Kitti test set (moderate difficulty) using the average precision metric (AP), higher is better.

**ACNet/ACNet+:** On Caltech, we use the enlarged training set and follow the training routine described in Section 2.3. We bootstrap several times until our small network gets saturated. For the ACNet/ACNet+ pipeline, we get a decent performance with and without involving the purging strategy from Section 2.5, see Figure 3. Moreover, when comparing to other lightweight methods, ACNet/ACNet+ are able to achieve a high detection rate (∼90%, ∼94%, respectively). As it can be seen in Figure 3, the underlying detectors ACF/ACF+ are outperformed by a large gap of ∼20 percent points (pp) in case of ACNet and ∼10 pp in case of ACNet+. As it will be discussed in the next Section, this improvement is made at a computationally very low price. On Kitti, the number of the pedestrians in the training set is much smaller than on Caltech, which makes it difficult to saturate the network. A comparison to other methods is shown in Figure 3. Note that on Kitti we do not use our purging module from Section 2.5. Overall, an improvement of ∼8 pp is made in case of ACNet.

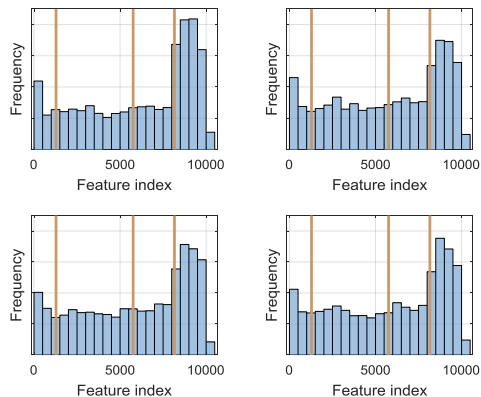

Figure 4. Selected feature set used for training ACNetBRF on Kitti training dataset. ACNet channels and ACF features are combined. The three vertical lines separate the respective feature types. The features are concatenated in the following order from left to right: ACF features, channels from convolution layers 1, 2, and 3. The channels are not aggregated or further processed. The top raw shows the feature distribution of the first (left) and second (right) 512 weak classifiers and the bottom raw shows the same distribution for the third (left) and fourth (right) 512 weak classifiers.

**ACNetBRF/ACNet+BRF:** As described in Section 2.4, we use the same bootstrapped training set which was collected during the ACNet/ACNet+ training to train the extensions ACNetBRF/ACNet+BRF, without using any additional data or further bootstrapping. As can be seen the trend is consistent showing that, when combining the features (CNN ∪ ACF) as described in Section 2.4, BRF has a better classification ability than the 2 fully connected layers of our CNNs. As can be seen in Figure 3, in both datasets we can improve through this extension which requires almost no additional computational effort, see also Figure 5. By having a closer look at the feature selection, it can be seen that the features coming from the first and the second layers are chosen almost equally often, see Figure 4. It seems that their discriminative power does not outperform the low level ACF feature representation. Features coming from third layer on the other hand, which have the largest receptive field and higher degree of abstraction, seem to be highly preferred. Specially in the beginning phase of the cascade (the first 512 weak classifiers), these features seem to be very effective in distinguishing the classes and dominate the feature selection.

**ACNet+BRFResNet:** We finally extend our ACNet+BRF pipeline with a pre-trained ResNet50 model. To saturate this deep model we continue bootstrapping, as mentioned in Section 2.4. Since this model is by far the costliest component in the pipeline, it is placed after NMS. The reached performance on both test sets can be seen in Figure 3 (ACNet+BRFResNet).

### 3.1 Comparison to the state-of-the-art methods

In Figure 5, we show a trade-off between speed and performance of the state-of-the-art methods on Caltech and Kitti datasets. As can be seen, ACNet and its extensions run at a low computation time when compared to other methods, even compared to those running on GPU.

**Caltech:** Our best result on Caltech (ACNet+BRFResNet) is very competitive to the currently leading method (RPN+BF [43], 9.6% MR). Clearly, leading methods rely on a fully CNN-based setup and use CNN methods for both collecting and evaluating proposals. Such methods have a large number of parameters and require expensive hardware (GPU), on which CNNs run multiple times faster, to obtain a sufficient run-time. As a pendant to this trend, our simple pipeline has demonstrated that even with little resources, a satisfying performance can be achieved.

**Kitti:** As mentioned before the pedestrian on Kitti are smaller (25 pixel) than on Caltech. This seems to make the detection task way more challenging and many of the well performing methods on Caltech (even the best performing method, RPN+BF) suffer from this effect (see Figure 5, compare left (Caltech) with right (Kitti)). The currently leading method on Kitti, MS-CNN [46], uses a multi-scale framework to overcome this difficulty. Such mechanism could also help our pipeline to improve in such cases, which would be an interesting topic for future work.

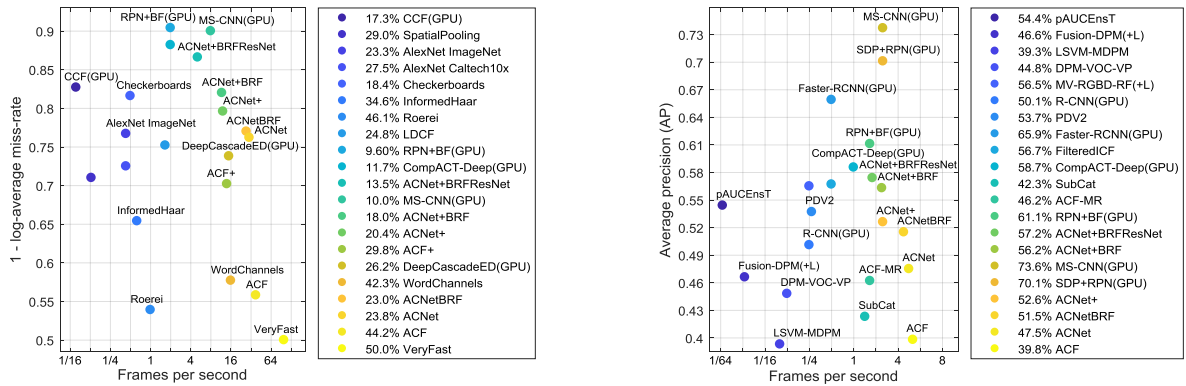

Figure 5. Performance of pedestrian detection methods in the accuracy-speed space. The higher the methods are located, the better the performance is. In the legend, the methods are ranked after their run-time. The right sphere (faster run-time) is mostly occupied by the different variants of our method, of which the detailed descriptions can be found in Section 2.

## 3.2 Run-time analysis

We adjust the cascade weights so that ACF and ACF+ run at an operating point, where both return, before NMS, an average of 40 candidates per image. At this operating point, the detectors run faster than at the default operating point. We are purging the proposed bounding boxes from the detector (this is only done on Caltech dataset) as described in Section 2.5 and re-score the remaining proposals with our presented ACNet/ACNet+ and apply NMS to the re-scored bounding boxes. At the chosen operating point, the run-time for ACNet and ACNet+ is 30 and 12 frames per seconds (FPS), respectively. All our experiments were conducted in serial mode on a modern desktop PC with an i5-4690 CPU. In Table 3, we decompose the run-time of the used modules. Note that, on Kitti the imagery has a resolution fluctuating around $1242 \times 375$ pixel and the pedestrians minimum height is 25 pixel (on Caltech we have $640 \times 480$ pixel images, with pedestrians above 50 pixel) which is why the methods run slower on the Kitti dataset. Figure 5, compares the performance of the proposed ACNet and ACF-related methods on Caltech and Kitti test sets. It can be seen that we achieve a solid performance gain compared to our underlying ACF and ACF+ proposal generators. These improvements are achieved on a very low computational price (see Table 3) and can be exploited exhaustively by making use of GPU computation.

|         | ACF   | ACNet | ACNetBRF | ACF+  | ACNet+ | ACNet+BRF | ACNet+BRFResNet |
|---------|-------|-------|----------|-------|--------|-----------|-----------------|
| Caltech | 0.026 | 0.033 | 0.036    | 0.070 | 0.083  | 0.085     | 0.196           |
| Kitti   | 0.156 | 0.177 | 0.208    | 0.371 | 0.398  | 0.413     | 0.555           |

Table 3. Run-time comparison of ACF/ACF+ and ACNet and its extensions on Caltech and Kitti datasets. Times are given in seconds per frame using a single CPU core.

## 4. CONCLUSION

In this work, we have presented a novel approach that conveniently combines one of the fastest sliding window detectors in the literature, ACF, with a small CNN. In particular, the pipeline is designed in a highly efficient way so that all of the heavy computation is done just once and shared between the detector and the CNN. By doing so, feature computation and resizing of the CNN input can be skipped. Furthermore, thanks to the enriched - via increasing the number of channels - and compressed - via pixel aggregation - CNN input, our pipeline is able to achieve a competitive performance in real-time, not to mention that the proposed pipeline could further benefit from the use of GPU. The simpler design and smaller model size of our CNN make it possible to be deployed in low-consumption hardware such as embedded chips. Most importantly, the fast run-time of the pipeline and its extensions allow using them as highly effective proposal generators. Future work includes investigating new, larger and more powerful architectures and exploiting scale and context, which are ongoing topics in the literature.